\documentclass[conference]{IEEEtran}

\usepackage{cite}
\usepackage{amsmath,amssymb,amsfonts}
\usepackage{dsfont}
\usepackage{algorithmic}
\usepackage{algorithm}
\usepackage{graphicx}
\usepackage{textcomp}
\usepackage{xcolor}
\usepackage{booktabs}
\usepackage{multirow}
\usepackage{tikz}
\usepackage{hyperref}
\usepackage{balance}
\usepackage{enumitem}

\newcommand{\system}{SkillSmith}

\newcommand{\stepnum}[1]{%
  \tikz[baseline=(char.base)]{
    \node[shape=circle,draw,inner sep=0.4pt,line width=0.35pt,
    font=\scriptsize] (char) {#1};
  }%
}

\begin{document}

\title{\system{}: Enhancing Locally Deployed Agents via Automatic Skill Construction and Evolution}

\author{
  \IEEEauthorblockN{Xinle Jiang\IEEEauthorrefmark{1}, Remy Xie\IEEEauthorrefmark{2}, and Ming Tang\IEEEauthorrefmark{1}\IEEEauthorrefmark{3}}
  \IEEEauthorblockA{\IEEEauthorrefmark{1}Southern University of Science and Technology, China}
  \IEEEauthorblockA{\IEEEauthorrefmark{2}AOE Tech Labs Ltd., Hong Kong}
  \IEEEauthorblockA{\IEEEauthorrefmark{3}Corresponding author}
}

\maketitle

% ============================================================
\begin{abstract}
LLM-based agent frameworks now act as personal assistants for multi-step tasks. Existing agent frameworks such as OpenClaw commonly follow the Cloud Agent depolyment mode using closed-source cloud LLMs as backbone model, which may expose private user information and incur repeated LLM-calling costs. Local Agents address these deployment concerns by depolying frontier open-source SLMs on user-controlled devices, but their task effectiveness still lags far behind Cloud Agents. Through diagnostic analysis, we reveal that the limited effectiveness of Local Agents with frontier SLM backbones mainly comes from missing environment knowledge caused by limited backbone model scale including environment rules and operation procedures. To supply such knowledge non-parametrically, context-efficiently, and without expert authoring, we present \system{}, a Cloud--Local Agent collaboration framework that uses Skill as a context-efficient knowledge carrier, automatic constructs Skill from Cloud Agent task exploration and evolves Skill using Local Agent execution feedback to enhance a frozen Local Agent. Experiments on daily agent task datasets AppWorld and WorkBench show that the automatically generated Skill enables the Local Agent with Qwen3.6-27B(SLM) to achieve task effectiveness comparable to Cloud Agents with frontier LLMs, outperform the strongest non-parametric baselines, reduce average actions per task from 36.1 to 9.9 on AppWorld-Normal, and generalize to other SLM backbone models without rerunning Skill construction.
\end{abstract}

%\begin{IEEEkeywords}
%Edge Intelligence, LLM Agent, Skill Evolution, Cloud-Edge Collaboration, Environment Knowledge
%\end{IEEEkeywords}

% ============================================================
\section{Introduction}
\label{sec:intro}
%!TEX root = ../main.tex
% =============================================================
% Section I: Introduction
% INFOCOM-style introduction structure, following Jupiter's rhetorical form:
% P1. Application trend and motivation for on-device deployment.
% P2. The central bottleneck that prevents practical local-agent deployment.
% P3. Three findings that localize the bottleneck to environment knowledge (§II-B).
% P4. Limitations of existing non-parametric methods, distilled into R1--R3 (§II-C).
% P5. SkillSmith positioning, with R1--R3 mapped to design choices (§II-D).
% P6. The two technical problems P1/P2 (§II-D).
% P7. Design overview: Offline Learning Phase and Online Serving Phase (§III-A).
% P8. Contributions.
% 
Large language model (LLM)-based agent frameworks such as
OpenClaw~\cite{openclaw2026docs}, Claude
Code~\cite{anthropic2026claudecode}, and
Hermes~\cite{nous2026hermesagent} now act as personal assistants for
multi-step tasks, including sending email, scheduling, and mobile-app operation. Most existing agent frameworks are deployed as cloud-hosted agents (Cloud Agents), which use closed-source cloud LLMs as backbone models. Since each execution step requires sending local task context to the cloud LLM, Cloud Agents may expose private user information and incur repeated LLM calling costs. To address this deployment overhead, researchers have proposed locally deployed agents (Local Agents)~\cite{jin2024cecollm,kang2025agentdistillation}, where a frontier open-source small language model (SLM) is deployed as the backbone model on the user-controlled local or edge device. Since the task context remains on device throughout execution, Local Agents reduce privacy exposure and avoid repeated cloud-LLM calling costs.

\begin{figure}[!t]
    \centering
    \includegraphics[width=\columnwidth]{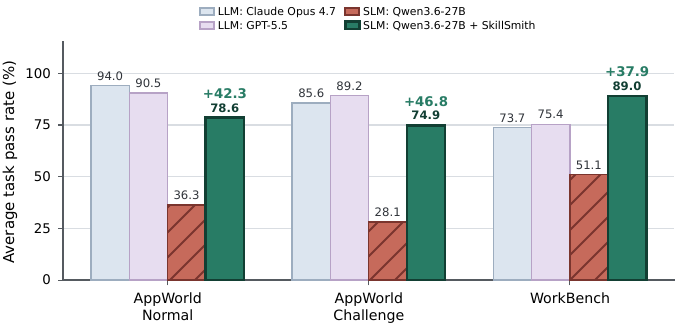}
    \caption{Task pass rates under three deployment settings: Cloud Agents with large LLM backbones, a Local Agent with an SLM backbone, and the same Local Agent augmented with \system{}. Green ``$+\Delta$'' labels denote the absolute
    gain of \system{} over the baseline Local Agent using the same
    Qwen3.6-27B backbone.}
    \label{fig:intro_pass_rates}
\end{figure}

However, we observe that Local Agents still lack real-world task effectiveness by using frontier open-source SLMs as the backbone model. As shown in Fig.~\ref{fig:intro_pass_rates}, the task pass rates of the Local Agent with Qwen3.6-27B are
only 36.3\%, 28.1\%, and 51.1\% in AppWorld-Normal, AppWorld-Challenge, and
WorkBench tasks, respectively, far below Cloud Agents with frontier LLMs. A natural explanation is the model scale of the backbone model. But this explanation is underspecified. As agent execution depends on at least two factors: core agentic capabilities (such as
tool use, instruction following and general reasoning), and environment
knowledge (including environment rules and operation procedures). This explanation does not clarify what concrete factors model scale causes. We therefore analyze this gap empirically.

Our analysis yields three findings that specify the model-scale explanation.
First, frontier open-source models such as Qwen3.6-27B are already comparable to closed-source LLMs on core agentic capabilities, suggesting that these capabilities are not primarily determined by parameter scale. Second, Local Agent failures
primarily stem from missing environment knowledge in the backbone model, including
environment rules and operation procedures that determine how the Agent should
act. This is where the effect of model scale: larger cloud LLMs can encode broader environment knowledge in their parameters, whereas a smaller local SLM may miss it. Third, preliminary experiments show that non-parametrically providing the missing environment knowledge such as injecting into prompt, can improved task effectiveness of Local Agent without fine-tuning the backbone model.

These findings motivate us to provide environment knowledge externally rather
than update the Local Agent's backbone model. However, prior methods that supply environment knowledge non-parametrically are not directly applicable to a context-limited Local Agent, which can be organized in three families.
\textit{Prompt-based} methods~\cite{brown2020fewshot,hsieh2023tooldoc} are the most direct: using system prompt as the carrier of environment knowledge to enhance Local Agent. However, constructing such prompts relies on expert prior knowledge and
repeated manual revision. Moreover, since only a small subset of the knowledge
is relevant to any single task, exposing the entire collection at every task
consumes the limited context of local SLMs, resulting in \emph{context
inefficiency}. \textit{Memory-based} methods~\cite{zhao2024expel,wang2024awm}
remove most of this manual effort by distilling trajectories and evaluator
feedback into reusable memories. However, the resulting memories are not always reliably environment knowledge due to noise in execution trajectories and distillation errors from models.And these methods lack a mechanism to verify and correct these memories. \textit{Skill-based} methods~\cite{anthropic2025skills} use Skill as a more suitable carrier than the prompt through \emph{progressive disclosure}: this makes the context cost depend on the knowledge needed by the current task, rather than on the size of the entire knowledge collection. Constructing a high-quality Skill, however, still relies on expert authoring and evolution. Recent work automates the subsequent evolution~\cite{ni2026trace2skill,alzubi2026evoskill,zhang2026coevoskills} through rollout feedback, but this process relies on initial usable Skill and therefore does not completely remove the expert dependency. These limitations suggest the lack of methods that can jointly satisfy three requirements for practical environment knowledge provision: (R1) a suitable carrier that avoids context inefficiency, (R2) automatic construction of high-quality initial knowledge without expert authoring, and (R3) automatic iterative evolution with verification and correction from Local Agent execution feedback.

% -------------------------------------------------------------
% P5. SkillSmith positioning: map R1--R3 to Skill carrier, offline Cloud Agent
%     assistance, and online local serving advantages.
% -------------------------------------------------------------
To satisfy R1 to R3 jointly, we present \system{}, a Cloud--Local Agent
collaboration framework that automatically constructs and evolves high-quality
Skills to enhance the frozen Local Agent. \system{} adopts the Skill as a context-efficient carrier for environment knowledge (R1), adotpts the Cloud Agent to replace human experts in initial Skill construction (R2), and adopts the Cloud Agent based on Local Agent execution feedback to drive iterative Skill evolution (R3). 

% -------------------------------------------------------------
% P6. Bridge question and technical problems.
% -------------------------------------------------------------
This design leverages
Cloud Agent's richer parametric environment knowledge and its stronger
reflection and summarization abilities. However, Cloud Agent assistance alone is insufficient to make this framework reliable. Realizing it raises two technical problems. \textbf{P1: How can the Cloud Agent construct a high-quality initial Skill from representative task executions and available task feedback?} Directly prompting the Cloud Agent to write a Skill is unreliable because task descriptions under-specify the actual
environment, so the generated Skill may hallucinate unsupported rules or miss
implicit procedures. \textbf{P2: How can the initial Skill be iteratively
evolved based on Local Agent's execution feedback?} Naively feeding the current Skill, failed trajectories, and evaluator feedback to the Cloud Agent is ineffective: failures are ambiguous, recurring defects are hard to identify across long traces, and unconstrained revisions can bloat the Skill or regress previously successful tasks.

Inspired by how human experts construct and evolve Skills from task trials and
execution feedback, \system{} addresses the two problems by designing a multi-agent collaboration mechanism with two stages. To
solve P1, the \emph{Skill Creation Stage} first performs \textit{Explore and
Reflect}: the Cloud Agent executes representative tasks and summarizes each trajectory into a Reflection Report. It then performs
\textit{Create Initial Skill}: the Cloud Agent clusters the reports, distills reusable environment knowledge, and merges it into a length-bounded initial Skill. To solve P2, the \emph{Skill Evolution Stage} first performs \textit{Execute and Compress}: the Local Agent executes tasks with the current Skill, and failed executions are compressed into Failure Reports. It then performs \textit{Evolve Skill}: the Cloud Agent clusters recurring failures, identifies Skill-addressable defects, and applies bounded revisions. 

In summary, this paper makes the following contributions:
\begin{itemize}
    \item We reveal that the limited task effectiveness of Local Agents with frontier SLM backbones mainly comes from missing environment knowledge caused by limited backbone model scale. We further show that such knowledge can be supplied non-parametrically to substantially improve Local Agent performance, avoiding fine-tuning the backbone model.
    \item To supply environment knowledge non-parametrically, context-efficiently, and without expert authoring, we present \system{}, a Cloud--Local Agent collaboration framework that automatically constructs and evolves high-quality Skills to enhance a frozen Local Agent.
    \item We evaluate \system{} on two daily agent task datasets, AppWorld and WorkBench. Experiments show that the generated Skill enables a Local Agent with Qwen3.6-27B to achieve task effectiveness comparable to Cloud Agents with frontier LLMs, outperforming the strongest non-parametric baseline on each dataset by 11.3 to 26.2 points, reducing average AppWorld-Normal actions per task from 36.1 to 9.9, and generalize to other SLM backbone models without rerunning Skill construction.
\end{itemize}

% ============================================================
\section{Background and Motivation}
\label{sec:background}
%!TEX root = ../main.tex
% =============================================================

% 本节按四步展开：§II-A 建立 Agent 架构与云-边部署形态；
% §II-B 论证 Local Agent 的差距是知识问题而非能力问题；
% §II-C 说明提供环境知识本身的三个子问题，然后引入Skill，并对其进行定义。说明其相比于提示词是一个更好地载体。但不确定是否要再次引入相关工作分析。
% $II-D 引入我们自己的工作作为过渡。

% ----------------------------------------------------------------
% Fig. 1 —— Agent 的 ReAct 架构与云/边部署形态。
% 图注: Agent Frameworks 和部署情况示意图。颜色图例对应 Observation/Thought/Action，红色字体展示用户隐私信息可能被上传到云端的情况。
% ----------------------------------------------------------------
\begin{figure}[t]
    \centering
    \includegraphics[width=\columnwidth]{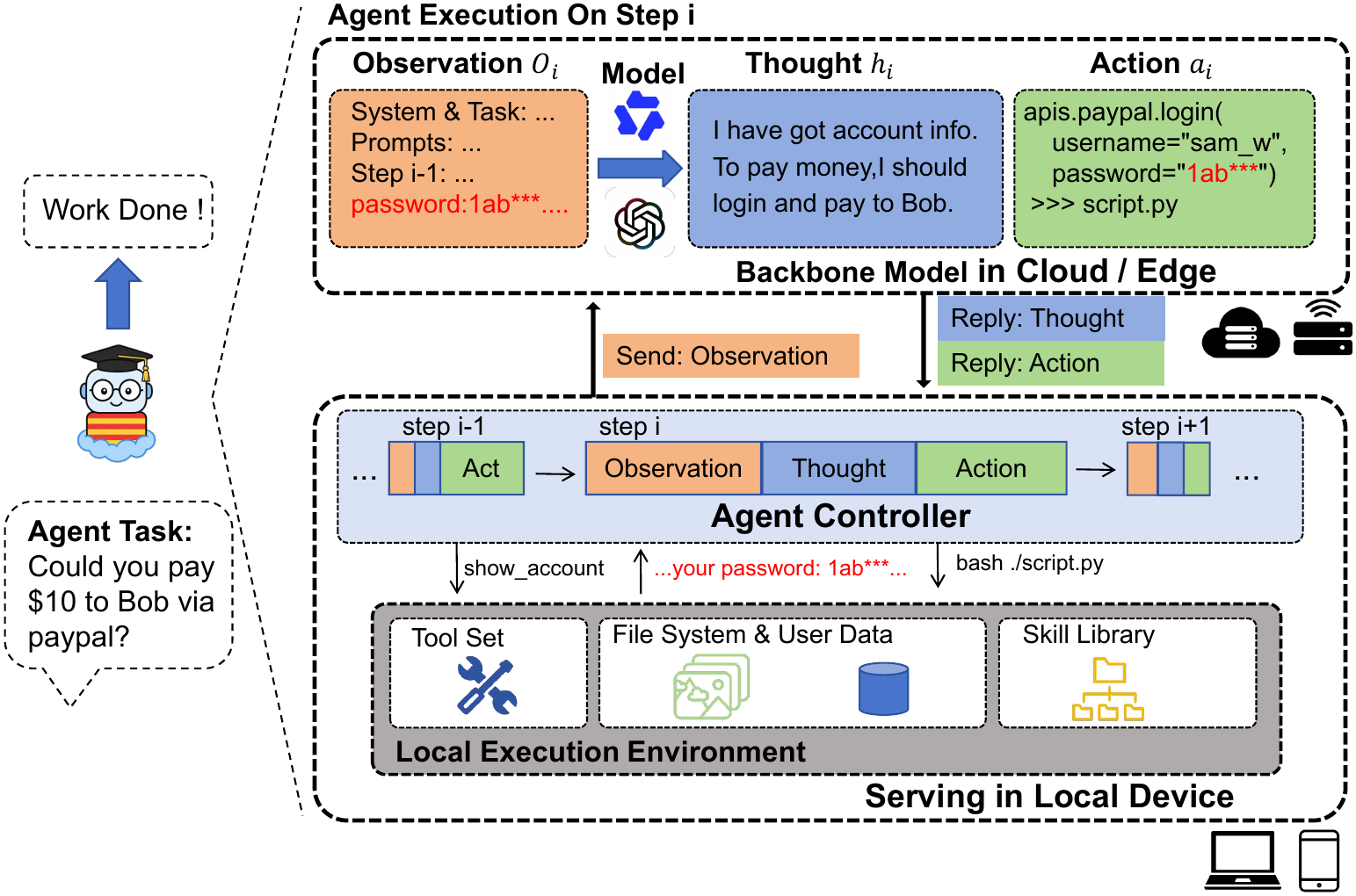}
    \caption{LLM-based agent framework and two deployment modes: Cloud Agent and Local Agent. Red
text marks private user information that may be included in the cumulative
observation and uploaded to the cloud.}
    \label{fig:agent-arch}
\end{figure}

% ================================================================
% §II-A  LLM-based Agent Frameworks and Deployment
% [Section Contract: II-A]
% 中文原文：
%   主流的 LLM-based Agent 框架（如 OpenClaw、Claude Code、Hermes）都由 Agent Controller、本地执行环境和 backbone model 三部分组成，并采用 ReAct 范式~\cite{yao2023react} 完成任务，每一步包含环境 observation、模型 thought 和 action。如图~\ref{fig:agent-arch} 所示，Controller 负责组织并推动这一循环：接收本地环境返回的 observation，维护累计 Observation，将其发送给 backbone model，并把模型返回的 action 交给本地环境执行。在第 $i$ 步，Controller 执行上一步动作 $a_{i-1}$ 后得到环境反馈 $o_i$，并更新累计 Observation $O_i = O_{i-1} + o_i$；backbone model 读取 $O_i$ 后返回 thought $h_i$ 和下一步 action $a_i$，Controller 在本地环境中执行 $a_i$，如此循环直至任务完成或 Agent 停止。
%   在这一共同框架下，Agent 有两种部署方式。两者都将 Controller 和本地执行环境部署在用户可控的设备上，区别在于 backbone model 的部署位置：cloud agent 将闭源 LLM 部署在远程云端服务器上，local agent 则将开源小语言模型（SLM）与 Controller 部署在同一用户设备上。由于 $O_i$ 可能包含任务信息、用户文件、账号凭据和应用状态，cloud agent 必须在每一步通过网络将其发送到远程模型，因而带来隐私暴露风险、重复 API 调用成本和网络时延；local agent 在设备内处理 $O_i$，使轨迹不离开本地，从而避免这些云端开销并保留本地执行的隐私与时延优势。
% ================================================================
\subsection{LLM-based Agent Framework and Two Deployment Modes}
\label{sec:background:agent-framework}

LLM-based agent frameworks, including OpenClaw~\cite{openclaw2026docs},
Claude Code~\cite{anthropic2026claudecode}, and
Hermes~\cite{nous2026hermesagent}, consist of three components: the Agent
Controller, the local execution environment, and the backbone model. Most of
these frameworks adopt the ReAct paradigm~\cite{yao2023react}, where execution
proceeds through repeated observation, thought, and action steps. As illustrated
in Fig.~\ref{fig:agent-arch}, the Controller maintains this process by
accumulating observations from the local execution environment, sending the
cumulative observation to the backbone model, and executing the returned action
locally. At step $i$, after action $a_{i-1}$ is executed, the environment returns
a new observation $o_i$. The Controller updates the cumulative observation as
$O_i=(O_{i-1}, o_i)$, where $(\cdot,\cdot)$ denotes appending the new
observation to the ordered context, and
sends $O_i$ to the backbone model. The backbone model then produces thought $h_i$ and next
action $a_i$, which the Controller executes in the local environment. This loop
continues until the task is completed or the agent stop.

Agents built on this framework have two deployment modes. The mainstream mode is the Cloud Agent, which keeps the Agent Controller and local execution environment
on the local device, but deploys a closed-source LLM backbone in the
cloud, often with more than one trillion parameters. In this mode, the Controller
sends the cumulative observation $O_i$ to the cloud model at every
step~\cite{openclaw2026docs,anthropic2026claudecode}. Since $O_i$ may contain task instructions, user-file contents and application state, this deployment can expose private user information to the
cloud while also incurring repeated API costs. To avoid these costs, researchers have proposed the Local Agent mode~\cite{jin2024cecollm,kang2025agentdistillation}, which deploys an open-source small language model
(SLM), typically below 35B parameters~\cite{qwen36}, as the backbone on the user-controlled
local or edge device. In this mode, $O_i$ remains on device throughout execution,
reducing privacy exposure and avoiding repeated cloud-model API calls.

% ================================================================
% §II-B  Agent执行任务场景的失败分析
% ================================================================

\subsection{Failure Analysis of Agent Task Execution}
\label{sec:background:capability-gap-isnt-the-cause}

\begin{table}[t]
    \centering
    \caption{Core agentic capability results for Qwen3.6-27B (SLM) and Claude
Opus 4.5 (LLM), reported in the Qwen3.6-27B technical report~\cite{qwen36}.}
    \label{tab:capability-parity}
    \setlength{\tabcolsep}{3.5pt}
    \begin{tabular}{lrrr}
        \toprule
        Benchmark & Qwen3.6-27B & Opus 4.5 & $\Delta$ (pp) \\
        \midrule
        SWE-bench Verified~\cite{jimenez2024swebench} & 77.2 & 80.9 & $-3.7$ \\
        Terminal-Bench 2.0~\cite{terminalbench} & 59.3 & 59.3 & $\phantom{-}0.0$ \\
        QwenClawBench~\cite{qwenclawbench} & 53.4 & 52.3 & $+1.1$ \\
        GPQA Diamond~\cite{rein2023gpqa} & 87.8 & 87.0 & $+0.8$ \\
        \bottomrule
    \end{tabular}
\end{table}

\begin{figure}[t]
    \centering
    \includegraphics[width=\columnwidth]{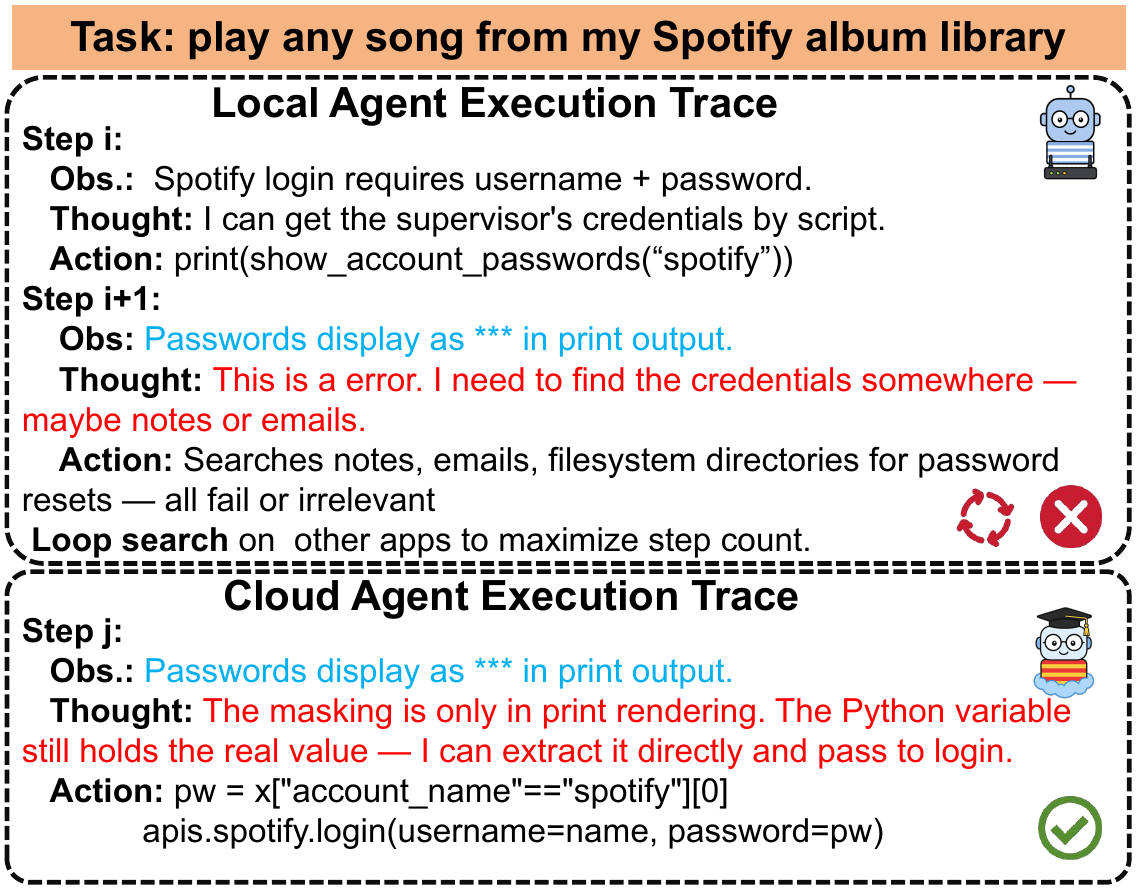}
    \caption{Representative failures caused by missing
    environment-specific procedures. The Local Agent
    treats a visually masked password as unavailable and searches
    elsewhere; the Cloud Agent recognizes that the credential remains
    accessible and succeeds.}
    \label{fig:failure-compare}
\end{figure}

\begin{table}[t]
    \centering
    \caption{resolution after injecting manually authored
environment knowledge. Reslv. after Knowl. denotes Resolved after injecting environment knowledge.}
    \label{tab:failure-attribution}
    \setlength{\tabcolsep}{3.5pt}
    \begin{tabular}{lrrr}
        \toprule
        Root Cause & Failures & Ratio & Reslv. after Knowl. \\
        \midrule
        Missing environment rule & 19 & 52.8\% & 15/19 \\
        Missing operation procedure & 14 & 38.9\% & 11/14 \\
        \textbf{Knowledge-related subtotal}
            & \textbf{33} & \textbf{91.7\%} & \textbf{26/33} \\
        \midrule
        Invalid action format & 1 & 2.8\% & 0/1 \\
        Reasoning or instruction error & 2 & 5.6\% & 1/2 \\
        \textbf{Capability-related subtotal}
            & \textbf{3} & \textbf{8.3\%} & \textbf{1/3} \\
        \midrule
        \textbf{Total}
            & \textbf{36} & \textbf{100.0\%} & \textbf{27/36} \\
        \bottomrule
    \end{tabular}
\end{table}

Despite the privacy protection and cost-saving benefits of local deployment,
Local Agents with frontier open-source SLMs still show limited task
effectiveness, falling far behind Cloud Agents on all three task sets
(Fig.~\ref{fig:intro_pass_rates}). A natural explanation is the difference in backbone model scale. However, this explanation is still underspecified, because it does not clarify what concrete execution gap is caused by model scale. We therefore conduct a deeper analysis.

\textbf{Finding 1: Frontier open-source SLMs already have core agentic
capabilities comparable to closed-source LLMs.}
Core agentic capabilities refer to the core abilities required for agent
execution, including tool use, instruction following, and general reasoning.
Table~\ref{tab:capability-parity} compares Qwen3.6-27B with Claude Opus
4.5~\cite{anthropic2026claudemodels} across benchmarks that cover these capabilities. SWE-bench Verified evaluates
software-level tool use; Terminal-Bench
evaluates command-line task execution and instruction following; QwenClawBench
provides an integrated evaluation of task execution; and GPQA Diamond
reflects advanced reasoning ability. Across these benchmarks, Qwen3.6-27B is
comparable to Claude Opus 4.5, suggesting that model scale does not
lead to a significant difference in core agentic capabilities.

\textbf{Finding 2: Local Agent failures primarily stem from missing environment knowledge in the backbone model.} Completing application tasks requires both core agentic capabilities and environment knowledge, including the rules and procedures that govern action selection. Since Finding 1 suggests that the Local Agent already has comparable core capabilities, the remaining gap is more likely due to missing environment knowledge in the backbone model. Larger cloud LLMs can encode more environment knowledge, whereas a smaller local SLM may miss such knowledge. This is reflected in failure trajectories: Fig.~\ref{fig:failure-compare} shows that the Local Agent misjudges a visually masked password as inaccessible, while the Cloud Agent applies the correct rule and succeeds. To quantify this effect, we randomly sampled 36 failure trajectories from Local Agent with Qwen3.6-27B on AppWorld-Normal. Table~\ref{tab:failure-attribution} shows that 33 failures (91.7\%) are caused by missing environment rules or operation procedures, indicating that the remaining effectiveness gap mainly comes from environment knowledge rather than core agentic capability.

\textbf{Finding 3: Providing environment knowledge non-parametrically
improves Local Agent task execution.} A common way to address missing environment knowledge is to collect additional data to fine-tune the backbone model~\cite{kang2025agentdistillation}. However, it is often impractical due to limited training data and computational resources, so we examine whether a non-parametric method is sufficient. We injected manually authored environment knowledge into the system prompt and reran the same 36 failed tasks. The guidance provides general missing rules and procedures rather than task-specific answers.Table~\ref{tab:failure-attribution} shows that this intervention resolves \textbf{26}/33 knowledge-related failures, indicating that application-specific environment knowledge can repair most such failures without updating model parameters.

% ================================================================
% §II-C  Agent Skill机制和自动化创建Skill的需求
% ================================================================
% 然而，现有 Skill 的创建和迭代仍主要依赖人类专家，难以跟上新的业务场景。专家通常先结合自己对应用场景知识和任务，编写包含主手册、附件及其调用条件的初始 Skill；随后让 Agent 搭载Skill在目标任务上反复执行，并根据失败轨迹和结果反馈判断Skill存在的问题并进行逐一人工修改。这个过程既需要对应用场景的丰富认知，也需要大量的时间投入和对失败情况的精准归因，但往往实际Local Agent部署者并不具备。II-B 的观察表明，Cloud Agent 更擅长从未知环境的反馈中反思和纠错，而Local Agent 的失败轨迹直接反映目标模型所缺少的知识。我们因此认为，Skill的创建与迭代可以由二者协作自动完成：利用 Cloud Agent 从执行轨迹中归纳环境知识，再更新 Local Agent 的失败反馈持续修正已有的Skill。
\subsection{Non-parametric Environment Knowledge Provision}
\label{sec:background:skill-mechanism}
These findings motivate us to provide environment knowledge by non-parametric methods rather than update the Local Agent’s back-
bone model. However, existing non-parametric methods for providing environment knowledge have significant limitations which can be roughly grouped into prompt-based, memory-based, and skill-based methods~\cite{jiang2026sok}.

\textit{Prompt-based methods}~\cite{brown2020fewshot,hsieh2023tooldoc} directly inject environment knowledge into the system prompt, but they rely on expert-written prompts and often overload the limited context of local SLMs with irrelevant knowledge. \textit{Memory-based methods}~\cite{zhao2024expel,wang2024awm,zhou2025memento,fang2025memp,liu2025structured} distill trajectories and feedback into reusable memories, but noisy executions and model-reflection errors can make these memories unreliable, and they usually lack mechanisms for verification and correction.

\textit{Skill-based methods} provide a more structured carrier compare to prompt. Following
Anthropic Agent Skills~\cite{anthropic2025skills}, we define an \emph{Agent
Skill} $\mathcal{S}$ as
\begin{equation}
    \mathcal{S}=(M,\mathcal{R},\mathcal{C}),
    \label{eq:skill-definition}
\end{equation}
where $M$ is the main Markdown file \texttt{SKILL.md} (hereafter \emph{main
manual}), containing metadata, general instructions, and pointers to references
in $\mathcal{R}$; $\mathcal{R}=\{r_j\}_{j=1}^{J}$ stores
reference files with environment rules and operation procedures, and
$\mathcal{C}=\{\kappa_j\}_{j=1}^{J}$ records the loading condition for each
reference. Skills support \emph{progressive disclosure}: During execution, the Agent first reads the general guidance in $M$ and loads a
reference file $r_j \in \mathcal{R}$ only when its loading condition
$\kappa_j \in \mathcal{C}$ is satisfied. This progressive disclosure makes Skills context-efficient for local SLMs because
the Agent places only task-relevant environment knowledge into the context,
rather than the entire knowledge collection. Recent methods such as Trace2Skill~\cite{ni2026trace2skill} and
EvoSkill~\cite{alzubi2026evoskill} further show
that Skills can be automatic evolved from Agent rollout feedback~\cite{wang2026skillx,xia2026skillrl,yang2026skillopt}, but they still assume a usable initial Skill remain expert-authored.

% ================================================================
% §II-D  Design Goal and Technical Challenges
% ================================================================
% 受到exprot从从所在应用场景的任务、agent执行轨迹中执行结果反馈中提炼有效提炼信息的启发+Cloud agent具有比较强的分析能力，为此我们提出了一个由 Cloud Agent 辅助、面向 Local Agent 的非参数化 Skill 自动构建与演化框架来致力于同时解决Q1-Q3的问题。
% 我们的系统从处理流程上可以分为\emph{Offline Learning Phase} 和 \emph{Online Serving Phase}。Offline Learning Phase 包含Skill Creation 和Skill Evolution 两个阶段。首先，\emph{Skill Creation Stage} Cloud Agent 会基于代表性训练任务 和 evaluator feedback，从中提炼可复用的环境知识并构造初始 Skill。随后，\emph{Skill Evolution Stage} Cloud agent 会与 Local Agent协作，根据local agent搭载当前 Skill 执行代表性任务的失败和未按预期调用Skill和它的reference的情况，修订 Skill 中尚未覆盖或未能触发使用条件的情况。最后得到Skill将会在Online Serving Phase 被Local Agent被一渐进披露的方式使用，从而有效提升task effictive。
% 实现这一框架面临两个关键挑战。 \textbf{C1: 如何从有限的 Cloud Agent 轨迹中构造可复用的初始 Skill？} Cloud Agent 的轨迹包含发现环境规则所需的操作经验，但也混杂了任务特定的参数、探索过程和偶然的执行选择。有限的代表性任务不能直接给出完整且无冲突的环境知识；将所有轨迹简单汇总，又会把偶然行为误写为通用规则，并使知识文档不断膨胀。更进一步，系统不仅需要决定“写入什么知识”，还需要决定哪些知识应始终放在主手册中，哪些知识只应在特定场景下通过 reference 加载，以及 Local Agent 如何识别相应的加载条件。因此，Skill Creation 本质上是从有限且混杂的执行经验中构造具有正确覆盖范围和层次结构的 Skill，而非一次普通的轨迹摘要。
% \textbf{C2: 如何从 Local Agent 的失败中可靠地演化 Skill？} Local Agent 的失败结果并不能直接说明 Skill 应如何修改。相同的 evaluator failure 可能由缺少环境知识、未加载相关 reference、已加载的指导不完整、Local Agent 未能遵循已有指导，或文本 Skill 无法弥补的模型能力限制所导致。若不区分这些原因而直接根据失败追加规则，Skill 会积累冗余甚至错误的知识，并可能干扰原本已经成功的任务。因而，Skill Evolution 需要结合 Local Agent 的执行轨迹与 evaluator feedback，识别其中能够由 Skill 修复的知识缺口，并对当前 Skill 进行受约束的局部演化。
\subsection{Design Goals and Technical Challenges}
\label{sec:background:design-goals}

These limitations suggest three requirements for practical non-parametric
environment knowledge provision: (R1) a suitable carrier that avoids context
inefficiency, (R2) automatic construction of high-quality initial environment
knowledge without expert authoring, and (R3) automatic iterative evolution from
Local Agent feedback. No existing unified framework is designed to satisfy all three requirements. To meet them, \system{} adopts Skills as the carrier for R1
and uses Cloud--Local Agent collaboration to replace human experts in Skill
construction and evolution for R2 and R3.

Realizing this design raises two technical problems. \textbf{P1:} How can the
Cloud Agent construct a high-quality initial Skill from representative task
executions and available task feedback? \textbf{P2: }How can the initial Skill be reliably evolved from Local Agent execution feedback?

% ============================================================
\section{System Design}
\label{sec:design}
%!TEX root = ../main.tex
This section presents \system{}, which enhances a frozen Local Agent by
automatically constructing and evolving Skills. We first overview the two-phase
workflow, then formulate Skill learning as black-box optimization, and finally
instantiate it through Skill Creation and Skill Evolution.

\begin{figure*}[t]
    \centering
    \includegraphics[width=\textwidth]{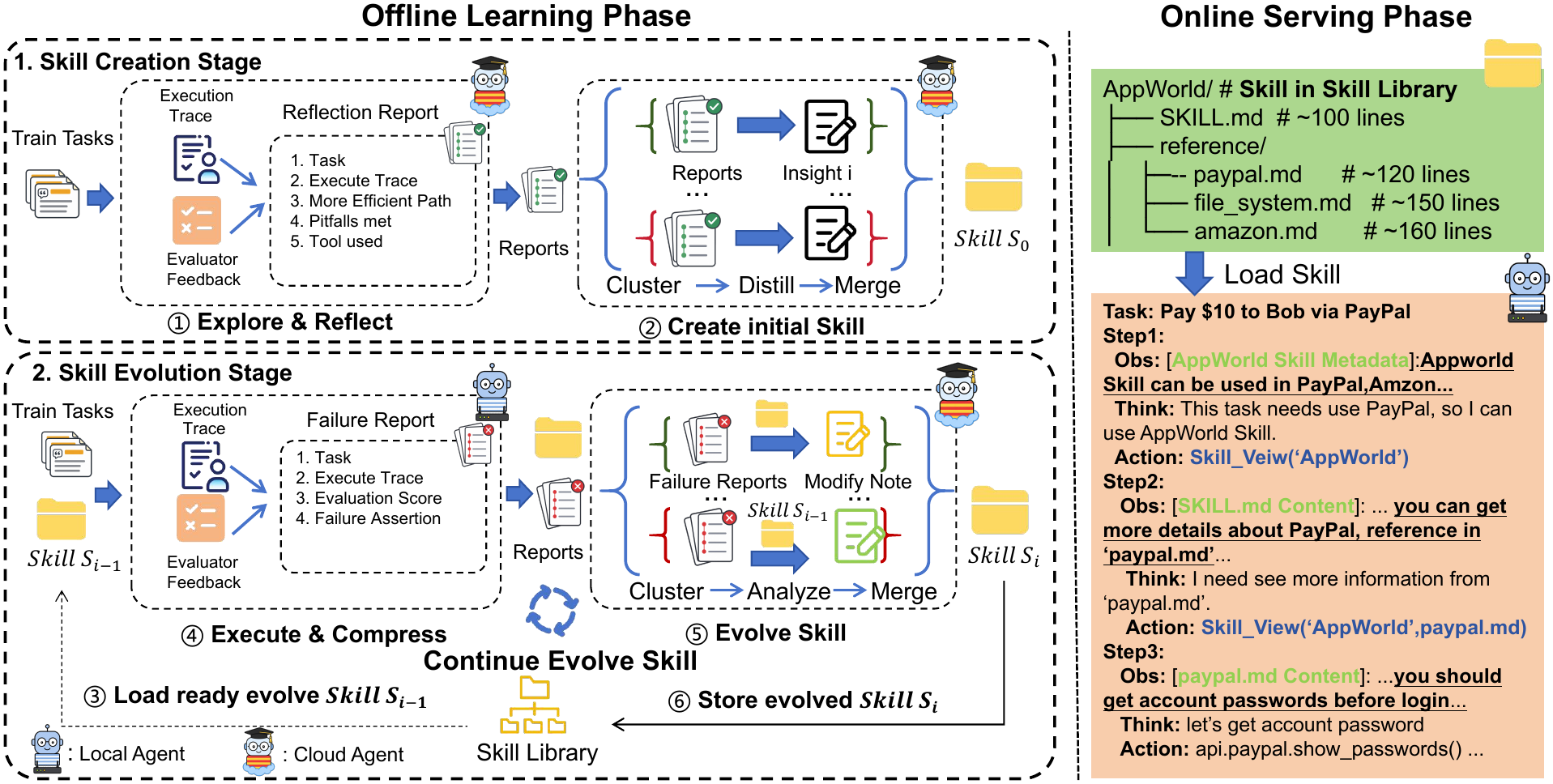}
    \caption{\system{} system overview.}
    \label{fig:architecture}
\end{figure*}

\subsection{System Overview}
\label{sec:design:overview}
% 这个系统应该包括两个phase。第一个phase，根据local agent应用场景自带的任务和评估器，在非人工干预下通过Local and Cloud Agent collaboration方式来得到高质量的Skill.第二个phase,local agent应用产出的Skill，有效提升任务effectivness。并且全程无需Cloud LLM，保护隐私。
\system{} has an \emph{Offline Learning Phase} and an \emph{Online Serving
Phase}, as shown in Fig.~\ref{fig:architecture}. Offline Learning Phase constructs
and evolves a target-scenario Skill from representative tasks and evaluator feedback through Cloud--Local Agent collaboration. The Online Serving Phase deploys the frozen Local Agent with the produced Skill to improve task effectiveness, with no Cloud LLM calls and all user data kept local.

Offline Learning Phase contains two connected stages. In the \emph{Skill Creation} stage, the inputs are training tasks sampled from the target application scenario and the user-provided evaluator. The Cloud Agent executes each task, obtains evaluator feedback, and writes Reflection Reports(\stepnum{1}). \system{} then clusters the Reflection Reports, distills reusable insights from each cluster, and merges them into the initial Skill $\mathcal{S}_0$ (\stepnum{2}).

The \emph{Skill Evolution} stage is an iterative loop. At iteration $i$,
\system{} loads the ready-to-evolve Skill $\mathcal{S}_{i-1}$ from the Skill Library (\stepnum{3}). The Local Agent executes the training tasks with $\mathcal{S}_{i-1}$, and failed executions are compressed with evaluator feedback into Failure Reports (\stepnum{4}). The
Cloud Agent clusters these reports, attributes failures against the current Skill, and merges the modifications into $\mathcal{S}_i$ (\stepnum{5}) Finally, $\mathcal{S}_i$ is stored back into the Skill Library for
the next iteration (\stepnum{6}).

\subsection{Problem Formulation}
\label{sec:design:formulation}

To clarify how the Offline Learning Phase obtains a high-quality Skill f, we cast Offline Learning as black-box Skill optimization. We first define
Skill-conditioned execution and the objective, then introduce Skill Creation and Skill Evolution as operators that search for an approximate solution.

Let $\mathcal{T}$ denote the task distribution of the target application
scenario, and let $\mathcal{D}_{\mathrm{train}}$ be a set of training tasks
sampled from $\mathcal{T}$ before deployment. Each task is
\begin{equation}
    \xi=(D,y^{*}),
\end{equation}
where $D$ is the task instruction and $y^{*}$ is the expected outcome used by
the evaluator. The Local Agent $G_s$ is frozen. Recall from
Sec.~\ref{sec:background:skill-mechanism} that a Skill is defined as
$\mathcal{S}=(M,\mathcal{R},\mathcal{C})$. We use $L_{\max}$ to denote the
Skill length budget imposed by the Local Agent's limited context capacity, and
let $\mathbb{S}_{L_{\max}}$ be the feasible space of valid Skills satisfying
this budget.

At each step $t$, let $H_t$ denote the interaction trajectory observed so far, including previous observations, thoughts, and actions. The Local Agent selects the next action based on the task instruction, the observed trajectory, and the loaded Skill:
\begin{equation}
    a_t \sim \pi_{G_s}\left(\cdot \mid D,H_t,\mathcal{S}\right).
\end{equation}

When the task terminates, the Local Agent produces an execution trace $\tau$ and
final output $\hat{y}$, which are evaluated against the expected outcome $y^{*}$:
\begin{equation}
    (r,e)=\operatorname{Eval}\left(\tau,\hat{y},y^{*}\right),
\end{equation}
where $r\in\{0,1\}$ denotes task success and $e$ is evaluator feedback. We write
$r(\mathcal{S};\xi)$ and $e(\mathcal{S};\xi)$ when the dependence on the Skill
$\mathcal{S}$ and the task $\xi$ needs to be made explicit.

The optimization objective is to find a valid Skill that maximizes the expected pass
rate of the frozen Local Agent:
\begin{equation}
    \mathcal{S}^{\star}
    =
    \arg\max_{\mathcal{S}\in\mathbb{S}_{L_{\max}}}
    \mathbb{E}_{\xi\sim\mathcal{T}}
    \left[r(\mathcal{S};\xi)\right],
\end{equation}
Since $\mathcal{T}$ is unknown, \system{} uses
the empirical pass rate on the sampled training tasks:
\begin{equation}
    J_{\mathcal{D}_{\mathrm{train}}}(\mathcal{S};G_s)
    =
    \frac{1}{|\mathcal{D}_{\mathrm{train}}|}
    \sum_{\xi\in\mathcal{D}_{\mathrm{train}}}
    r(\mathcal{S};\xi).
\end{equation}
This objective cannot be solved directly because $\mathbb{S}_{L_{\max}}$ is a
discrete natural-language document space and the Local Agent's pass rate has no
gradient. Inspired by TextGrad~\cite{yuksekgonul2024textgrad}, it treats LLM-generated feedback as a gradient-like signal and uses an LLM as a text optimizer to approach . Analogously,
\system{} uses the Cloud Agent $G_c$ as the Skill optimizer. It first constructs
an initial Skill from training tasks:
\begin{equation}
    \mathcal{S}_0
    =
    \operatorname{Create}
    (G_c,\mathcal{D}_{\mathrm{train}}).
\end{equation}
It then evolves the Skill for $R$ rounds, using Local Agent executions as
feedback and the Cloud Agent to propose updates:
\begin{equation}
    \mathcal{S}_i
    =
    \operatorname{Evolve}
    (G_c,G_s,\mathcal{S}_{i-1},\mathcal{D}_{\mathrm{train}}),
    \quad i=1,\ldots,R.
\end{equation}
The final deployed Skill is the version with the best empirical pass rate on
$\mathcal{D}_{\mathrm{train}}$.

%!TEX root = ../main.tex

\subsection{Skill Creation}
\label{sec:design:creation}
\begin{figure}[t]
    \centering
    \includegraphics[width=\columnwidth]{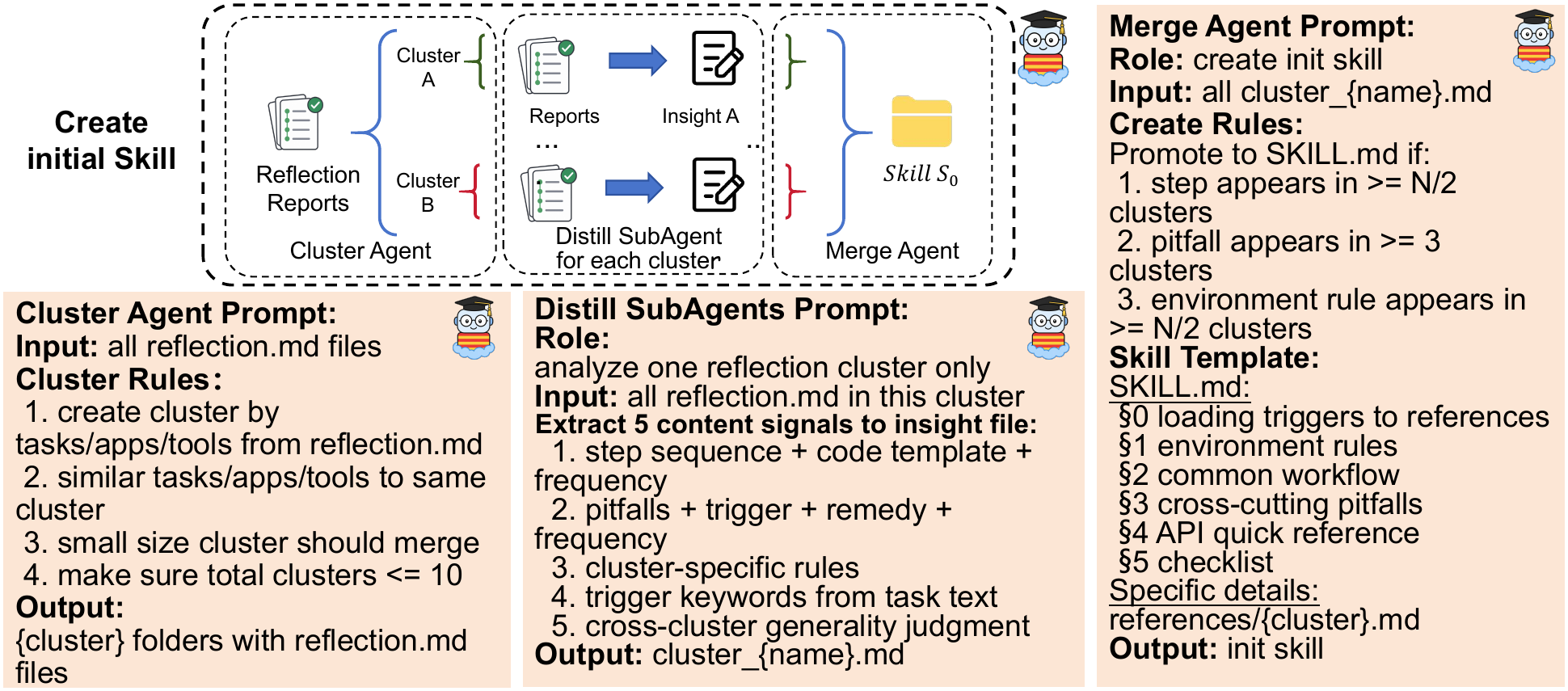}
    \caption{Skill Creation pipeline and prompts for three cloud-agent roles.}
    \label{fig:skill-creation}
\end{figure}
Skill Creation aims to produce a high-quality initial Skill
$\mathcal{S}_0$ without expert-authored initialization. A naive approach is to
ask the Cloud Agent $G_c$ to directly write a Skill based on the task description. This is unreliable because $G_c$ is not initially familiar with the target application scenario and may invent unsupported rules or miss implicit procedures. Inspired by how human experts learn a new application scenario and write Skills by trying tasks, checking outcomes, reflecting, and summarizing reusable insights, we design the
agentic Skill Creation pipeline, which consists of two steps: Explore and Reflect, and Create Initial Skill.

\textit{Explore and Reflect.} This step aims to collect material for Skill writing by letting the Cloud Agent execute target-scenario tasks and observe evaluator feedback. However, raw trajectories are not directly usable as environment knowledge: they contain redundant observations, exploratory detours, and failed attempts, while useful Skill material should be reusable experience such as procedures, environment rules, tool-use conventions, and common pitfalls. Inspired by prior work showing
that LLM agents can acquire reusable experience through self-reflection after
interaction~\cite{wang2023voyager,zhao2024expel,chen2024automanual}, \system{} uses the Cloud Agent to preprocess the sampled training tasks into a compact Reflection Report set:
$\mathcal{Q}\gets\textsc{ExecuteReflect}(G_c,\mathcal{D}_{\mathrm{train}})$.
This operator follows Algorithm~\ref{alg:offline-skill}: for each training task, the Cloud Agent executes the task, invokes the evaluator, and reflects on the execution trace and feedback. The Cloud Agent is prompted with a report template covering the task, a compressed execution trace, the evaluation result, a more efficient solution path, pitfalls encountered, and tools or resources used. The output is a report set
$\mathcal{Q}=\{q_i\}_{i=1}^{n}$, where
$n=|\mathcal{D}_{\mathrm{train}}|$. 

\textit{Create Initial Skill.}
This step aims to transform task-level Reflection Reports into a scenario-level
initial Skill $\mathcal{S}_0$. The main difficulty is scale and mixture: the reports often exceed 100 files and 10K lines in total, and they mix repeated procedures with task-specific details. Directly placing them into the Local Agent context would exceed the context budget and dilute attention. \system{} therefore creates the initial Skill through three cloud-agent roles, as illustrated in Fig.~\ref{fig:skill-creation}. (i) The Cluster Agent first groups the reports as
$\{\mathcal{Q}_c\}_{c=1}^{C}\gets\textsc{Cluster}(G_c,\mathcal{Q})$. The Cluster Agent groups reports by task type, application, and tool usage, merges small or semantically similar groups, and keeps the number of clusters $C$ bounded. (ii) A Distill SubAgent is assigned to each cluster
$\mathcal{Q}_c$ and extracts reusable signals into an insight file
$I_c=\textsc{Distill}(G_c,\mathcal{Q}_c)$, including step sequences, code or
API templates, pitfalls and remedies, cluster-specific rules, trigger keywords,
and judgments about whether an item is cluster-specific or cross-cluster.
(iii) The Merge Agent induces the initial Skill by
$\mathcal{S}_0\gets
\textsc{Merge}(G_c,\{I_c\}_{c=1}^{C},L_{\max})$: frequent cross-cluster rules and workflows are promoted to the
main manual, while scenario-specific details are written into reference files
with corresponding loading triggers, subject to the budget $L_{\max}$.

%!TEX root = ../main.tex

\subsection{Skill Evolution}
\label{sec:design:evolution}
Although Skill Creation provides an initial Skill $\mathcal{S}_0$, this Skill
is distilled mainly from Cloud Agent executions and may not fully match the
failure modes of the frozen Local Agent. Some knowledge may be missing, some
references may not be loaded under the right conditions, and some instructions
may be too weak for the Local Agent to follow reliably. Skill Evolution
therefore iteratively improves the current Skill $\mathcal{S}_{i-1}$ using
Local Agent failures, evaluator feedback, and Cloud Agent analysis. Each evolution iteration consists of two steps: Execute \& Compress and Evolve Skill.

\begin{figure}[t]
    \centering
    \includegraphics[width=\columnwidth]{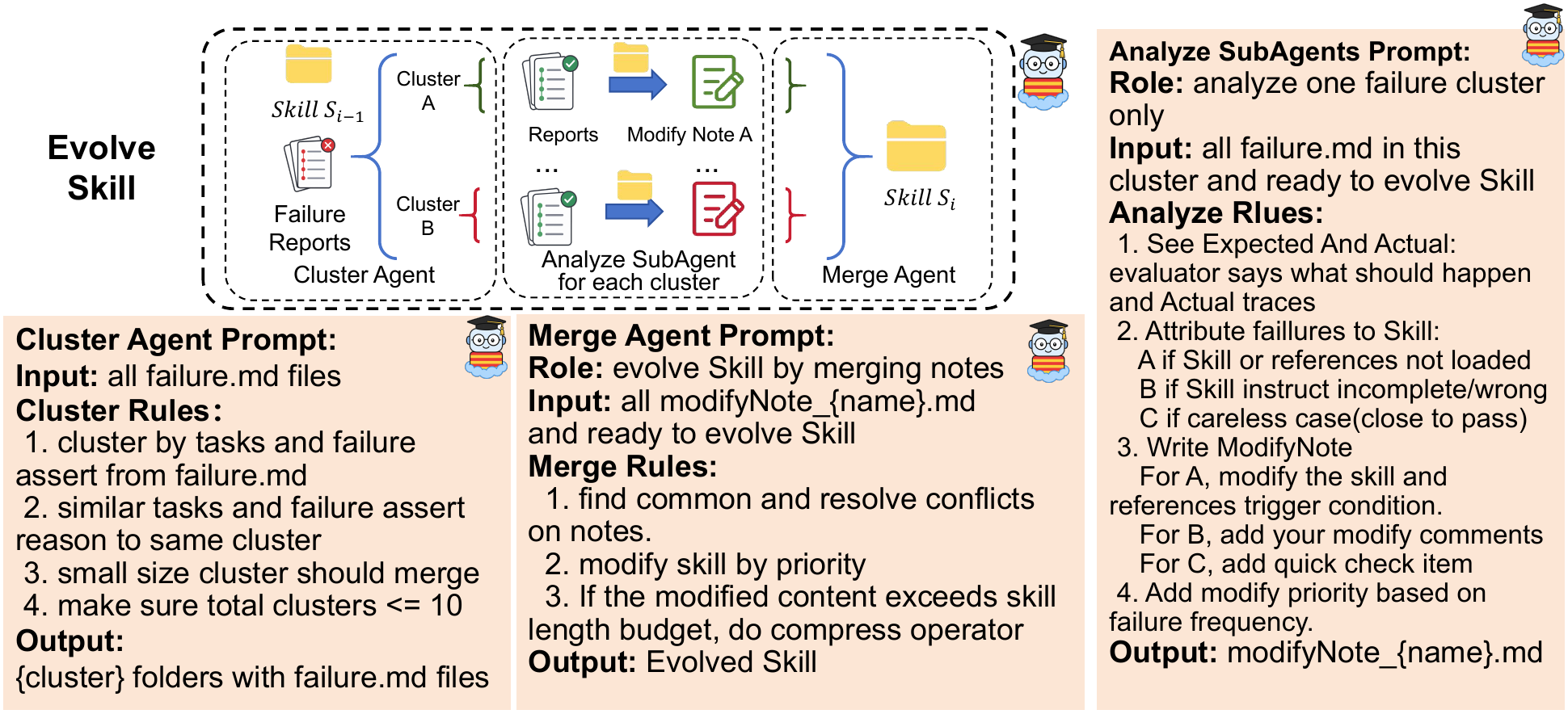}
    \caption{Skill Evolution pipeline and prompts for three cloud-agent roles.}
    \label{fig:skill-evolution}
\end{figure}

\textit{Execute and Compress.}
This step aims to expose what the current Skill still fails to support. The output is a failure-report set
$\mathcal{B}^{-}_{i}$, together with the current pass
rate $j_{i-1}$. At iteration $i$, \system{} runs
$(\mathcal{B}^{-}_{i},j_{i-1})\gets
\textsc{ExecuteCompress}
(G_s,G_c,\mathcal{S}_{i-1},\mathcal{D}_{\mathrm{train}})$. For each failed task, \system{} combines
the execution trace, evaluator feedback, and Skill-loading record into a compact Failure Report. This compression is necessary because raw failure traces are
long and noisy: they include observations, tool outputs, actions, and loaded
Skill content, while Skill editing only needs the critical mismatch between the
expected and actual execution.

\textit{Evolve Skill.}
This step aims to convert recurring Local Agent failures into length-bounded Skill
updates. The budget $L_{\max}$ is necessary because Local Agents using SLMs have tighter context constraints, and oversized Skill files make relevant instructions harder to attend to. The key difficulty is failure attribution: a failed task may indicate that the relevant Skill or reference was not loaded, that the loaded
instruction is incomplete or wrong, that the Local Agent made a near-miss
careless error, or that the failure is beyond textual repair. Blindly appending
every failure case would bloat the Skill and may damage tasks that already
succeed. Inspired by the workflow of expert Skill evolution, \system{} makes revisions reliable by first clustering recurring failures,
then attributing each cluster against the current Skill, and finally applying only
bounded edits with explicit priority. As shown in Fig.~\ref{fig:skill-evolution}, Local Agent evolves the Skill through three cloud-agent roles. (i) The Cluster Agent groups Failure Reports as $\{\mathcal{B}^{-}_{i,c}\}_{c=1}^{C_i}\gets
\textsc{Cluster}(G_c,\mathcal{B}^{-}_{i})$ by task type and failure assertion,
merges small or similar groups, and keeps the number of clusters $C_i$ bounded.
(ii) Each Analyze SubAgent inspects one failure cluster with the
ready-to-evolve Skill, $m_{i,c}\gets
\textsc{Analyze}(G_c,\mathcal{B}^{-}_{i,c},\mathcal{S}_{i-1})$.
It compares the evaluator's expected outcome with the actual trace, attributes
the failure to Skill loading, Skill content, near-miss execution, or
non-Skill-addressable causes, and writes a Modify Note with edit priority.
(iii) The Merge Agent combines all Modify Notes into the evolved Skill,
$\mathcal{S}_{i}\gets
\textsc{Merge}(G_c,\{m_{i,c}\}_{c=1}^{C_i},\mathcal{S}_{i-1},L_{\max})$.
Merge resolves conflicts, applies edits by priority, and compresses the Skill
when it exceeds the length budget. After each iteration, the evolved Skill $\mathcal{S}_{i}$ is evaluated by re-running \textsc{ExecuteCompress} to get failure feedback for evolution. After $R$ evolution iterations, \system{} selects the Skill version with the best empirical pass rate as
$\mathcal{S}^{\star}$. 

The full offline Skill optimization procedure is
summarized in Algorithm~\ref{alg:offline-skill}. The prompt examples in Fig.~\ref{fig:skill-creation} and Fig.~\ref{fig:skill-evolution} illustrate one implementation of the pipeline; They are not fixed components and can be adapted to user requirements or application-specific constraints.

% The algorithm mirrors the two offline stages.
\begin{algorithm}[!t]
\caption{Offline Skill Optimization in \system{}}
\label{alg:offline-skill}
\footnotesize
\begin{algorithmic}[1]
\REQUIRE Sampled training tasks $\mathcal{D}_{\mathrm{train}}$,
Cloud Agent $G_c$, Local Agent $G_s$, Evolution rounds $R$, Capacity $L_{\max}$
\ENSURE Deployed Skill $\mathcal{S}^{\star}$
\STATE \textbf{Part I: Skill Creation}
\STATE $\mathcal{Q}\gets\textsc{ExecuteReflect}
       (G_c,\mathcal{D}_{\mathrm{train}})$
\STATE $\{\mathcal{Q}_c\}_{c=1}^{C}\gets\textsc{Cluster}(G_c,\mathcal{Q})$
\STATE $I_c\gets\textsc{Distill}(G_c,\mathcal{Q}_c)$ for all $c$ in parallel
\STATE $\mathcal{S}_{0}\gets\textsc{Merge}
       (G_c,\{I_c\}_{c=1}^{C},L_{\max})$
\STATE \textbf{Part II: Skill Evolution}
\STATE $(\mathcal{B}^{-}_1,j_0)\gets\textsc{ExecuteCompress}
       (G_s,G_c,\mathcal{S}_{0},\mathcal{D}_{\mathrm{train}})$
\STATE $\mathcal{S}^{\star}\gets\mathcal{S}_{0}$; $j^{\star}\gets j_0$
\FOR{$i=1$ \TO $R$}
    \IF{$\mathcal{B}^{-}_i=\varnothing$}
        \STATE \textbf{break}
    \ENDIF
    \STATE $\{\mathcal{B}^{-}_{i,c}\}_{c=1}^{C_i}\gets
           \textsc{Cluster}(G_c,\mathcal{B}^{-}_i)$
    \STATE $m_{i,c}\gets\textsc{Analyze}
           (G_c,\mathcal{B}^{-}_{i,c},\mathcal{S}_{i-1})$
           for all $c$ in parallel
\STATE $\mathcal{S}_{i}\gets\textsc{Merge}
       (G_c,\{m_{i,c}\}_{c=1}^{C_i},\mathcal{S}_{i-1},L_{\max})$
    \STATE $(\mathcal{B}^{-}_{i+1},j_i)\gets\textsc{ExecuteCompress}
           (G_s,G_c,\mathcal{S}_{i},\mathcal{D}_{\mathrm{train}})$
    \IF{$j_i>j^{\star}$}
        \STATE $\mathcal{S}^{\star}\gets\mathcal{S}_{i}$;
               $j^{\star}\gets j_i$
    \ENDIF
\ENDFOR
\RETURN $\mathcal{S}^{\star}$
\end{algorithmic}
\end{algorithm}

% ============================================================
\section{Implementation and Evaluation}
\label{sec:eval}
%!TEX root = ../main.tex

Our experiments answer four research questions. 
\textbf{RQ1:} How effective is \system{} compared with state-of-the-art methods? 
\textbf{RQ2:} How much does \system{} cost to generate and use skills? 
\textbf{RQ3:} What are the contributions of each component?
\textbf{RQ4:} How well do the generated skills generalize across different foundation models? 

\subsection{Experimental Setup}
\label{sec:eval:setup}

\textbf{Datasets.}
We evaluate SkillSmith on two agentic datasets that cover daily-life and office-work scenarios. \textit{AppWorld}~\cite{trivedi2024appworld} requires agents to complete user instructions by interacting with simulated mobile applications via APIs, such as Amazon and Gmail, to sending emails, shopping and other daily activities. AppWorld contains two difficulty levels: AppWorld-Normal and AppWorld-Challenge, where the latter contains more cross-application tasks. \textit{WorkBench}~\cite{styles2024workbench} evaluates agents on workplace tasks involving calendar, analytics, project-management, and multi-domain activities. We randomly sample 180 AppWorld tasks (about 25\%) and 207 WorkBench tasks (about 30\%) as training set; the remaining tasks are used as test set for evaluation.

\textbf{Baselines.}
We compare \system{} with cloud agents powered by closed-source frontier LLMs and three families of non-parametric enhancement methods. 
\begin{itemize}
    \item \textbf{Closed-source frontier LLMs.}
    We adopt GPT-5.5~\cite{openai2026models} and Claude Opus 4.7~\cite{anthropic2026claudemodels} as backbone models for cloud agents.
    \item \textbf{Prompt-based methods.}
    These methods inject environment knowledge into the prompt: Few-shot ICL~\cite{brown2020fewshot} injects a small number of successful execution trajectories as in-context demonstrations, Doc-Prompting~\cite{hsieh2023tooldoc} injects API-level tool documentation and operation manuals.
    \item \textbf{Memory-based methods.}
    These methods store reusable experience extracted from agent trajectories and retrieve relevant memories during task execution: ExpeL~\cite{zhao2024expel} extracts natural-language insights and compact successful trajectories, AWM~\cite{wang2024awm} abstracts successful trajectories into workflow memories.
    \item \textbf{Skill-based methods.}
     Trace2Skill~\cite{ni2026trace2skill} evolves an existing skill by automatic revising skill based on execution traces. In our implementation, We initialize Trace2Skill with a initial skill derived from the description of datasets.
\end{itemize}

\textbf{Implementation details.}
We implement the offline Cloud Agent with the Claude Code SDK~\cite{anthropic2026claudecode} using Claude Opus 4.7~\cite{anthropic2026claudemodels} as the backbone model, and implement the Local Agent based on the Hermes agent framework~\cite{nous2026hermesagent} using Qwen3.6-27B~\cite{qwen36} as the backbone model. Skill Evolution runs for at  five rounds. We set $L_{\max}=400$ lines for the main manual and cap each reference file at 250 lines. For a fair comparison, all non-cloud methods are evaluated with the same Hermes agent framework and Qwen3.6-27B backbone model. Cloud-agent baselines also use the Hermes agent framework, but replace the local backbone with the corresponding cloud LLM. 
\subsection{Task Effectiveness (RQ1)}
\label{sec:eval:main-results}

\begin{table}[!t]
\centering
\caption{Task effectiveness on App-Nor.(AppWorld-Normal), App-Chal.(AppWorld-Challenge), and WB.(WorkBench); est Local Agent
results.}
\label{tab:main-results}
\begingroup
\setlength{\tabcolsep}{0pt}
\renewcommand{\arraystretch}{1.0}
\footnotesize
\begin{tabular*}{\columnwidth}{@{\extracolsep{\fill}}lccccc@{}}
\toprule
& \multicolumn{2}{c}{\textbf{App-Nor.}} & \multicolumn{2}{c}{\textbf{App-Chal.}} & \textbf{WB} \\
\cmidrule(lr){2-3}\cmidrule(lr){4-5}\cmidrule(l){6-6}
\textbf{Method} & \textbf{PR} & \textbf{AS} & \textbf{PR} & \textbf{AS} & \textbf{PR} \\
\midrule
\multicolumn{6}{l}{\textit{Closed-source frontier LLMs}} \\
Claude Opus 4.7~\cite{anthropic2026claudemodels} & 94.0 & 0.966 & 85.6 & 0.939 & 73.7 \\
GPT-5.5~\cite{openai2026models}                  & 90.5 & 0.960 & 89.2 & 0.962 & 75.4 \\
\midrule
\multicolumn{6}{l}{\textit{Vanilla with frontier SLMs}} \\
Qwen3.6-27B~\cite{qwen36} & 36.3 & 0.525 & 28.1 & 0.424 & 51.1 \\
\midrule
\multicolumn{6}{l}{\textit{Prompt-based methods}} \\
Few-shot ICL~\cite{brown2020fewshot}      & \underline{67.3} & 0.779 & 41.7 & 0.528 & 48.9 \\
Doc-Prompting~\cite{hsieh2023tooldoc}     & 28.0 & 0.508 & 21.6 & 0.451 & 47.4 \\
\midrule
\multicolumn{6}{l}{\textit{Memory-based methods}} \\
ExpeL~\cite{zhao2024expel}   & 65.9 & \underline{0.816} & \underline{62.1} & \underline{0.867} & 49.5 \\
AWM~\cite{wang2024awm}       & 51.5 & 0.749 & 45.6 & 0.630 & 48.7 \\
\midrule
\multicolumn{6}{l}{\textit{Skill-based methods}} \\
Trace2Skill~\cite{ni2026trace2skill} & 59.8 & 0.754 & 42.9 & 0.621 & \underline{62.8} \\
\system{} (ours) & \textbf{78.6} & \textbf{0.907} & \textbf{74.9} & \textbf{0.922} & \textbf{89.0} \\
\bottomrule
\end{tabular*}
\endgroup
\vspace{-6pt}
\end{table}

Table~\ref{tab:main-results} reports the main task-effectiveness results on AppWorld and WorkBench. We use pass rate (PR) as the primary metric for both datasets and additionally report the AppWorld average evaluator score (AS).

\textbf{Comparison with non-parametric enhancement methods.}
\system{} consistently achieves the best performance among all non-parametric enhancement methods. \system{} outperforms the strongest baseline on every dataset by at least 11.3
points, reaching 78.6\%, 74.9\%, and 89.0\% PR on AppWorld-Normal,
AppWorld-Challenge, and WorkBench, respectively. These results suggest that \system{} extracts higher-quality environment knowledge from training tasks and organizes it into Skill to enhance Local Agent. We also observe that prompt-based and memory-based methods bring little improvement and even slightly degrade performance on WorkBench. This suggests that, in some scenarios, useful environment knowledge cannot be obtained from one-shot context injection or memory extraction alone; instead, it requires multiple rounds of refinement based on execution feedback.

\textbf{Comparison with Cloud Agents.}
\system{} substantially narrows the effectiveness gap between the Local Agent and Cloud Agents. Compared with the strongest Cloud Agent on each dataset, the vanilla Local Agent is 57.7, 61.1, and 24.3 points lower in PR on AppWorld-Normal, AppWorld-Challenge, and WorkBench, respectively. After loading the \system{} Skill, the first two gaps narrow to 15.4 and 14.3 points, and the WorkBench result becomes a 13.6-point advantage over the strongest Cloud Agent. These results suggest that the Skill generated by \system{} substantially reduces the environment-knowledge deficit caused by the smaller backbone model, allowing the Local Agent to approach the task effectiveness of Cloud Agent and to exceed it some scenarios.

\subsection{Offline Cost and Online Overhead (RQ2)}
\label{sec:eval:cost}

\begin{figure}[!t]
\centering
\includegraphics[width=\columnwidth]{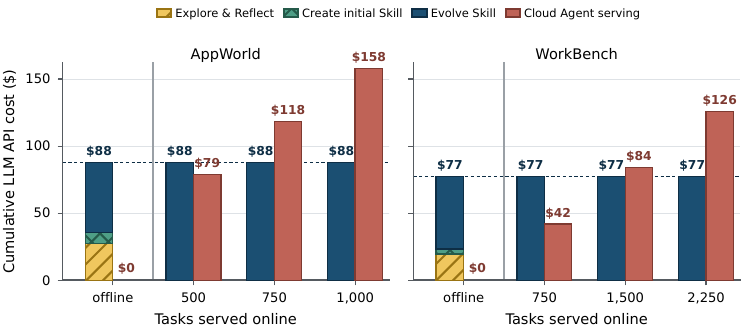}
\caption{LLM API cost of \system{} versus a Cloud Agent with Claude Opus 4.7.
Dashed lines denote \system{}'s one-time offline cost under local serving.}
\label{fig:offline-cost}
\vspace{-6pt}
\end{figure}

\textbf{Offline cost.}
Fig.~\ref{fig:offline-cost} compares the LLM API costs of \system{} and a Cloud Agent using Opus 4.7 as the backbone model, with costs broken down by phase. All numbers are collected from backend LLM API usage logs. For \system{}, the total cost of producing a deployable Skill is \$87.95 on AppWorld and \$77.27 on WorkBench, of which Skill Evolution accounts for \$52.40 and \$53.89, respectively. This is a one-time offline cost paid before deployment. By contrast, the Cloud Agent incurs LLM API cost for every served task, so its cumulative serving cost keeps increasing with usage. In this way, \system{} shifts cost from online serving to offline construction: after roughly 560 AppWorld tasks and 1{,}400 WorkBench tasks, the Cloud Agent has already spent more on serving than \system{} spent on building the Skill, while \system{} continues serving locally without additional LLM API cost.

\begin{table}[!t]
\centering
\caption{Online Qwen3.6-27B inference with and without the \system{} Skill.
Actions is the average number of actions per task. Cumulative context and
output tokens are summed over all inference steps of a task, not the size of a
single context window; both are reported in k tokens.}
\label{tab:online-overhead}
\setlength{\tabcolsep}{3.5pt}
\resizebox{\columnwidth}{!}{%
\begin{tabular}{lcccccc}
\toprule
& \multicolumn{3}{c}{\textbf{AppWorld-Normal}} & \multicolumn{3}{c}{\textbf{WorkBench}} \\
\cmidrule(lr){2-4}\cmidrule(l){5-7}
\textbf{Setting} & \textbf{Actions} & \textbf{Cum. Ctx.} & \textbf{Output} & \textbf{Actions} & \textbf{Cum. Ctx.} & \textbf{Output} \\
\midrule
Without Skill        & 36.1 & 450.7 & 6.3 & 8.0 & 32.6 & 2.0 \\
With \system{} Skill & 9.9  & 200.6 & 3.7 & 6.5 & 99.8 & 2.0 \\
\bottomrule
\end{tabular}}
\vspace{-6pt}
\end{table}

\textbf{Online inference overhead.}
Table~\ref{tab:online-overhead} compares the average number of actions per task
and the cumulative token overhead per task with and without the Skill generated
by \system{}. Loading a Skill may introduce additional context overhead as the
Local Agent needs to read the relevant procedural knowledge during execution.
However, this overhead can be partially mitigated, as the environment knowledge
in the Skill helps the agent avoid unnecessary trial-and-error and skip invalid
actions. On AppWorld, the no-Skill agent spends many actions on exploration and
failed attempts; after introducing the Skill, the agent completes tasks more
directly, reducing actions per task from 36.1 to 9.9, cumulative context tokens
from 450.7k to 200.6k, and output tokens from 6.3k to 3.7k.

\subsection{Ablation Study (RQ3)}
\label{sec:eval:ablation}

\begin{table}[!t]
\centering
\caption{Stage-level ablation. Values are PR (\%). Creation-only uses the
initial Skill before Skill Evolution.}
\label{tab:stage-ablation}
\setlength{\tabcolsep}{3.2pt}
\resizebox{\columnwidth}{!}{%
\begin{tabular}{lccc}
\toprule
\textbf{Setting} & \textbf{AppWorld-Normal} & \textbf{AppWorld-Challenge} & \textbf{WorkBench} \\
\midrule
No Skill & 36.3 & 28.1 & 51.1 \\
Skill Creation only & 66.3 & 71.6 & 87.8 \\
Creation + Evolution & \textbf{78.6} & \textbf{74.9} & \textbf{89.0} \\
\midrule
Creation gain & +30.0 & +43.5 & +36.7 \\
Evolution gain & +12.3 & +3.3 & +1.2 \\
\bottomrule
\end{tabular}}
\vspace{-6pt}
\end{table}

Table~\ref{tab:stage-ablation} shows ablation study of the two stages of \system{}: Skill Creation Stage adds 30.0 points on AppWorld-Normal, 43.5 points on AppWorld-Challenge, and 36.7 points on WorkBench over the no-Skill Local Agent. Skill Evolution Stage then adds 12.3, 3.3, and 1.2 points, respectively. These results show that Skill Creation is the dominant contributor, while Skill Evolution provides additional refinement.

\begin{figure}[!t]
\centering
\includegraphics[width=\columnwidth]{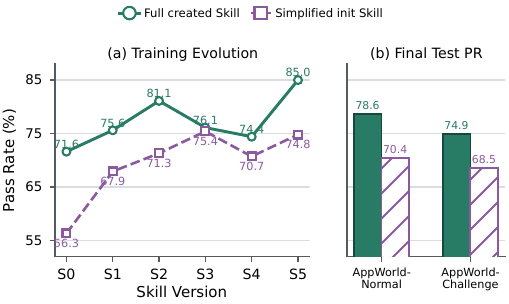}
\caption{Skill Evolution dynamics on AppWorld. (a) Training pass rate across
Skill versions on the training set. (b) Final pass rate on the test set.}
\label{fig:evolution-dynamics}
\vspace{-6pt}
\end{figure}

Figure~\ref{fig:evolution-dynamics} compares two different S0 initializations:
the full Skill produced by Skill Creation and a simplified init Skill that keeps
only the main procedural body of the full version. We then apply five rounds of Skill Evolution to both initializations. With the full created Skill, training PR starts at 71.6\% and reaches 85.0\% at S5; with the simplified init Skill, training PR starts at 56.3\% and reaches 75.4\% at S3 before oscillating. This explains why Evolution brings a smaller marginal gain in the full pipeline: Skill Creation already provides a strong initial solution, leaving less room for subsequent local refinement. Accordingly, \system{} supports two operating modes: Creation-only and Creation + Evolution.

\subsection{Skill Transferability (RQ4)}
\label{sec:eval:transfer}

\begin{table}[!t]
\centering
\caption{Cross-backbone Skill transferability. The Qwen3.6-27B Skill is loaded
unchanged by each deployment backbone. Values are PR (\%).}
\label{tab:transfer}
\setlength{\tabcolsep}{2.7pt}
\footnotesize
\begin{tabular}{lcccccc}
\toprule
& \multicolumn{3}{c}{\textbf{AppWorld-Normal}} & \multicolumn{3}{c}{\textbf{WorkBench}} \\
\cmidrule(lr){2-4}\cmidrule(l){5-7}
\textbf{Backbone} & \textbf{No Skill} & \textbf{+Skill} & \textbf{$\Delta$}
                  & \textbf{No Skill} & \textbf{+Skill} & \textbf{$\Delta$} \\
\midrule
Qwen3.6-27B (self) & 36.3 & 78.6 & +42.3 & 51.1 & 89.0 & +37.9 \\
Qwen3.6-30B-A3B~\cite{qwen330ba3b} &  3.1 & 23.2 & +20.1 & 29.4 & 60.2 & +30.8 \\
Gemma-4-31B~\cite{google2026gemma431b} & 17.9 & 66.1 & +48.2 & 73.9 & 81.7 & +7.8 \\
\bottomrule
\end{tabular}
\vspace{-6pt}
\end{table}

This part studies whether a Skill generated for \system{} can still be useful when the backbone model changes. In this experiment, we keep the same agent framework and generated Skill unchanged, and only replace the Local Agent backbone model. Table~\ref{tab:transfer} shows consistent gains across all evaluated backbones. On the original Qwen3.6-27B backbone, the Skill improves PR by 42.3 points on AppWorld-Normal and 37.9 points on WorkBench. Without rerunning offline construction, the same Skill also improves Qwen3.6-30B-A3B by 20.1 and 30.8 points, and Gemma-4-31B by 48.2 and 7.8 points, respectively. Overall, these results suggest that the Skills generated by \system{} exhibit strong cross-backbone transferability.

% ============================================================
\section{Related Work}
\label{sec:related}
%!TEX root = ../main.tex

\subsection{Large--Small Model Collaboration}
\label{sec:related:cloud-assisted}

Large and small models are often combined through online selection or offline
parameter transfer. Routing and cascade systems such as
FrugalGPT~\cite{chen2024frugalgpt} choose among models per request, while
cloud--edge methods such as CE-CoLLM~\cite{jin2024cecollm} keep a local model as
the default executor and invoke a cloud model for difficult requests. These
methods still keep the cloud model in the serving loop. Agent
Distillation~\cite{kang2025agentdistillation} moves the transfer offline by
training small agents from cloud-agent trajectories, but requires local-model
updates. In contrast, \system{} uses a Cloud Agent only before deployment to
build a non-parametric Skill; the deployed Local Agent remains frozen and runs
without cloud API calls.

\subsection{Knowledge enhancement and Skill Evolution}
\label{sec:related:agent-skills}
These work equip agents with reusable non-parametric knowledge.
Prompt-based methods~\cite{brown2020fewshot,hsieh2023tooldoc}
place demonstrations or manuals in context. For memory-based 
methods, Memento~\cite{zhou2025memento} learns procedural memory from interaction so agent behavior can be adapted without fine-tuning LLM weights. REMe~\cite{cao2026reme} maintains dynamic procedural memory and refines it with new experience and feedback. Memp~\cite{fang2025memp} explores how agent procedural memory can be organized for future reuse. MACLA~\cite{forouzandeh2026macla} builds hierarchical procedural memory with Bayesian selection and contrastive refinement.

For skill-based methods, EvoSkill~\cite{alzubi2026evoskill} refiness skills from execution trajectories, SkillRL~\cite{xia2026skillrl} distills past experience into a hierarchical SkillBank and recursively evolves it with RL, SkillOpt~\cite{yang2026skillopt} optimizes skill documents through bounded edits and validation gates, and CoEvoSkills~\cite{zhang2026coevoskills} uses co-evolutionary verification to improve skills and reduce regressions.

% ============================================================
\section{Conclusion}
\label{sec:conclusion}
%!TEX root = ../main.tex

This paper introduces \system{}, a Cloud--Local Agent collaboration framework that automatically constructs and evolves high-quality Skills to enhance a frozen Local Agent. Our experiments show that the generated Skill enables a Local Agent with Qwen3.6-27B to achieve effectiveness comparable to Cloud Agents, outperforming the strongest non-parametric baseline on each dataset by 11.3 to 26.2 points, reducing average actions per task, and generalize to other SLM backbone models without rerunning Skill construction.

% ============================================================
\balance
\bibliographystyle{IEEEtran}
\bibliography{references}

@inproceedings{yao2023react,
  title={{ReAct}: Synergizing Reasoning and Acting in Language Models},
  author={Yao, Shunyu and Zhao, Jeffrey and Yu, Dian and Du, Nan and Shafran, Izhak and Narasimhan, Karthik and Cao, Yuan},
  booktitle={International Conference on Learning Representations (ICLR)},
  year={2023}
}

@article{wang2023voyager,
  title={Voyager: An Open-Ended Embodied Agent with Large Language Models},
  author={Wang, Guanzhi and Xie, Yuqi and Jiang, Yunfan and Mandlekar, Ajay and Xiao, Chaowei and Zhu, Yuke and Fan, Linxi and Anandkumar, Anima},
  journal={Transactions on Machine Learning Research},
  year={2024}
}

@article{yuksekgonul2024textgrad,
  title={Optimizing Generative {AI} by Backpropagating Language Model Feedback},
  author={Yuksekgonul, Mert and Bianchi, Federico and Boen, Joseph and Liu, Sheng and Lu, Pan and Huang, Zhi and Guestrin, Carlos and Zou, James},
  journal={Nature},
  volume={639},
  number={8055},
  pages={609--616},
  year={2025},
  doi={10.1038/s41586-025-08661-4}
}

@misc{anthropic2025skills,
  title={Equipping Agents for the Real World with Agent Skills},
  author={{Anthropic}},
  year={2025},
  howpublished={\url{https://www.anthropic.com/engineering/equipping-agents-for-the-real-world-with-agent-skills}},
  note={Announced October 16, 2025}
}

@inproceedings{trivedi2024appworld,
  title={{AppWorld}: A Controllable World of Apps and People for Benchmarking Interactive Coding Agents},
  author={Trivedi, Harsh and Khot, Tushar and Hartmann, Mareike and Manku, Ruskin and Dong, Vinty and Li, Edward and Gupta, Shashank and Sabharwal, Ashish and Balasubramanian, Niranjan},
  booktitle={Proceedings of the 62nd Annual Meeting of the Association for Computational Linguistics (ACL)},
  pages={16022--16076},
  year={2024}
}

@inproceedings{styles2024workbench,
  title={{WorkBench}: A Benchmark Dataset for Agents in a Realistic Workplace Setting},
  author={Styles, Olly and Miller, Sam and Cerda-Mardini, Patricio and Guha, Tanaya and Sanchez, Victor and Vidgen, Bertie},
  booktitle={Conference on Language Modeling (COLM)},
  year={2024}
}

@inproceedings{jimenez2024swebench,
  title={{SWE-bench}: Can Language Models Resolve Real-World {GitHub} Issues?},
  author={Jimenez, Carlos E. and Yang, John and Wettig, Alexander and Yao, Shunyu and Pei, Kexin and Press, Ofir and Narasimhan, Karthik},
  booktitle={International Conference on Learning Representations (ICLR)},
  year={2024}
}

@inproceedings{rein2023gpqa,
  title={{GPQA}: A Graduate-Level Google-Proof {Q\&A} Benchmark},
  author={Rein, David and Hou, Betty Li and Stickland, Asa Cooper and Petty, Jackson and Pang, Richard Yuanzhe and Dirani, Julien and Michael, Julian and Bowman, Samuel R.},
  booktitle={Conference on Language Modeling (COLM)},
  year={2024}
}

@article{terminalbench,
  title={Terminal-Bench: Benchmarking Agents on Hard, Realistic Tasks in Command Line Interfaces},
  author={Merrill, Mike A. and Shaw, Alexander G. and Carlini, Nicholas and Li, Boxuan and Raj, Harsh and Bercovich, Ivan and others},
  journal={arXiv preprint arXiv:2601.11868},
  year={2026}
}

@misc{qwenclawbench,
  title={{QwenClawBench}: A Real-User-Distribution Benchmark for Evaluating {OpenClaw} Agents},
  author={{Skylenage AI}},
  year={2026},
  howpublished={\url{https://huggingface.co/datasets/skylenage-ai/QwenClawBench}},
  note={Hugging Face Datasets}
}

@inproceedings{brown2020fewshot,
  title={Language Models are Few-Shot Learners},
  author={Brown, Tom B. and Mann, Benjamin and Ryder, Nick and Subbiah, Melanie and Kaplan, Jared and Dhariwal, Prafulla and Neelakantan, Arvind and Shyam, Pranav and Sastry, Girish and Askell, Amanda and Agarwal, Sandhini and Herbert-Voss, Ariel and Krueger, Gretchen and Henighan, Tom and Child, Rewon and Ramesh, Aditya and Ziegler, Daniel M. and Wu, Jeffrey and Winter, Clemens and Hesse, Christopher and Chen, Mark and Sigler, Eric and Litwin, Mateusz and Gray, Scott and Chess, Benjamin and Clark, Jack and Berner, Christopher and McCandlish, Sam and Radford, Alec and Sutskever, Ilya and Amodei, Dario},
  booktitle={Advances in Neural Information Processing Systems (NeurIPS)},
  volume={33},
  pages={1877--1901},
  year={2020}
}

@article{hsieh2023tooldoc,
  title={Tool Documentation Enables Zero-Shot Tool-Usage with Large Language Models},
  author={Hsieh, Cheng-Yu and Chen, Si-An and Li, Chun-Liang and Fujii, Yasuhisa and Ratner, Alexander and Lee, Chen-Yu and Krishna, Ranjay and Pfister, Tomas},
  journal={arXiv preprint arXiv:2308.00675},
  year={2023}
}

@inproceedings{zhao2024expel,
  title={{ExpeL}: {LLM} Agents Are Experiential Learners},
  author={Zhao, Andrew and Huang, Daniel and Xu, Quentin and Lin, Matthieu and Liu, Yong-Jin and Huang, Gao},
  booktitle={Proceedings of the AAAI Conference on Artificial Intelligence},
  volume={38},
  number={17},
  pages={19632--19642},
  year={2024}
}

@inproceedings{wang2024awm,
  title={Agent Workflow Memory},
  author={Wang, Zora Zhiruo and Mao, Jiayuan and Fried, Daniel and Neubig, Graham},
  booktitle={International Conference on Machine Learning (ICML), PMLR 267},
  pages={63897--63911},
  year={2025}
}

@inproceedings{cao2026reme,
  title={Remember Me, Refine Me: A Dynamic Procedural Memory Framework for Experience-Driven Agent Evolution},
  author={Cao, Zouying and Deng, Jiaji and Yu, Li and Zhou, Weikang and Liu, Zhaoyang and Ding, Bolin and Zhao, Hai},
  booktitle={Findings of the Association for Computational Linguistics: ACL 2026},
  pages={16803--16822},
  year={2026}
}

@article{zhou2025memento,
  title={Memento: Fine-tuning {LLM} Agents without Fine-tuning {LLMs}},
  author={Zhou, Huichi and Chen, Yihang and Guo, Siyuan and Yan, Xue and Lee, Kin Hei and Wang, Zihan and Lee, Ka Yiu and Zhang, Guchun and Shao, Kun and Yang, Linyi and Wang, Jun},
  journal={arXiv preprint arXiv:2508.16153},
  year={2025}
}

@inproceedings{fang2025memp,
  title={Memp: Exploring Agent Procedural Memory},
  author={Fang, Runnan and Liang, Yuan and Wang, Xiaobin and Wu, Jialong and Qiao, Shuofei and Xie, Pengjun and Huang, Fei and Chen, Huajun and Zhang, Ningyu},
  booktitle={Findings of the Association for Computational Linguistics: ACL 2026},
  pages={17490--17502},
  year={2026}
}

@inproceedings{forouzandeh2026macla,
  title={Learning Hierarchical Procedural Memory for {LLM} Agents through {Bayesian} Selection and Contrastive Refinement},
  author={Forouzandeh, Saman and Peng, Wei and Moradi, Parham and Yu, Xinghuo and Jalili, Mahdi},
  booktitle={Proceedings of the International Conference on Autonomous Agents and Multiagent Systems (AAMAS)},
  year={2026}
}

@article{ni2026trace2skill,
  title={Trace2Skill: Distill Trajectory-Local Lessons into Transferable Agent Skills},
  author={Ni, Jingwei and Liu, Yihao and Liu, Xinpeng and Sun, Yutao and Zhou, Mengyu and Cheng, Pengyu and Wang, Dexin and Zhao, Erchao and Jiang, Xiaoxi and Jiang, Guanjun},
  journal={arXiv preprint arXiv:2603.25158},
  year={2026}
}

@article{wang2026skillx,
  title={{SkillX}: Automatically Constructing Skill Knowledge Bases for Agents},
  author={Wang, Chenxi and Yu, Zhuoyun and Xie, Xin and Yao, Wuguannan and Fang, Runnan and Qiao, Shuofei and Cao, Kexin and Zheng, Guozhou and Qi, Xiang and Zhang, Peng and Deng, Shumin},
  journal={arXiv preprint arXiv:2604.04804},
  year={2026}
}

@article{xia2026skillrl,
  title={{SkillRL}: Evolving Agents via Recursive Skill-Augmented Reinforcement Learning},
  author={Xia, Peng and Chen, Jianwen and Wang, Hanyang and Liu, Jiaqi and Zeng, Kaide and Wang, Yu and Han, Siwei and Zhou, Yiyang and Zhao, Xujiang and Chen, Haifeng and Zheng, Zeyu and Xie, Cihang and Yao, Huaxiu},
  journal={arXiv preprint arXiv:2602.08234},
  year={2026}
}

@article{yang2026skillopt,
  title={{SkillOpt}: Executive Strategy for Self-Evolving Agent Skills},
  author={Yang, Yifan and Gong, Ziyang and Huang, Weiquan and Yang, Qihao and Zhou, Ziwei and Huang, Zisu and Li, Yan and Gao, Xuemei and Dai, Qi and Liu, Bei and Qiu, Kai and Yang, Yuqing and Chen, Dongdong and Yang, Xue and Luo, Chong},
  journal={arXiv preprint arXiv:2605.23904},
  year={2026}
}

@article{alzubi2026evoskill,
  title={{EvoSkill}: Automated Skill Discovery for Multi-Agent Systems},
  author={Alzubi, Salaheddin and Provenzano, Noah and Bingham, Jaydon and Chen, Weiyuan and Vu, Tu},
  journal={arXiv preprint arXiv:2603.02766},
  year={2026}
}

@article{zhang2026coevoskills,
  title={{CoEvoSkills}: Self-Evolving Agent Skills via Co-Evolutionary Verification},
  author={Zhang, Hanrong and Fan, Shicheng and Zou, Henry Peng and Chen, Yankai and Wang, Zhenting and Zhou, Jiayu and Li, Chengze and Huang, Wei-Chieh and Yao, Yifei and Zheng, Kening and Liu, Xue and Li, Xiaoxiao and Yu, Philip S.},
  journal={arXiv preprint arXiv:2604.01687},
  year={2026}
}

@article{kang2025agentdistillation,
  title={Distilling {LLM} Agent into Small Models with Retrieval and Code Tools},
  author={Kang, Minki and Jeong, Jongwon and Lee, Seanie and Cho, Jaewoong and Hwang, Sung Ju},
  journal={arXiv preprint arXiv:2505.17612},
  year={2025}
}

@article{chen2024frugalgpt,
  title={{FrugalGPT}: How to Use Large Language Models While Reducing Cost and Improving Performance},
  author={Chen, Lingjiao and Zaharia, Matei and Zou, James},
  journal={Transactions on Machine Learning Research},
  year={2024}
}

@inproceedings{chen2024automanual,
  title={{AutoManual}: Constructing Instruction Manuals by {LLM} Agents via Interactive Environmental Learning},
  author={Chen, Minghao and Li, Yihang and Yang, Yanting and Yu, Shiyu and Lin, Binbin and He, Xiaofei},
  booktitle={Advances in Neural Information Processing Systems (NeurIPS)},
  year={2024}
}

@misc{openclaw2026docs,
  title={{OpenClaw} Documentation},
  author={{OpenClaw}},
  year={2026},
  howpublished={\url{https://docs.openclaw.ai/}},
  note={Accessed July 22, 2026}
}

@misc{anthropic2026claudecode,
  title={{Claude Code} Documentation},
  author={{Anthropic}},
  year={2026},
  howpublished={\url{https://code.claude.com/docs/en/overview}},
  note={Accessed July 22, 2026}
}

@misc{anthropic2026claudemodels,
  title={{Claude} Model Overview},
  author={{Anthropic}},
  year={2026},
  howpublished={\url{https://platform.claude.com/docs/en/docs/about-claude/models/overview}},
  note={Accessed July 22, 2026}
}

@misc{openai2026models,
  title={{OpenAI} Model Documentation},
  author={{OpenAI}},
  year={2026},
  howpublished={\url{https://platform.openai.com/docs/models}},
  note={Accessed July 22, 2026}
}

@misc{nous2026hermesagent,
  title={{Hermes Agent}},
  author={{Nous Research}},
  year={2026},
  howpublished={\url{https://github.com/NousResearch/hermes-agent}},
  note={GitHub repository. Accessed July 22, 2026}
}

@article{jin2024cecollm,
  title={{CE-CoLLM}: Efficient and Adaptive Large Language Models Through Cloud-Edge Collaboration},
  author={Jin, Hongpeng and Wu, Yanzhao},
  journal={arXiv preprint arXiv:2411.02829},
  year={2024}
}

@misc{qwen36,
  title={{Qwen3.6-27B} Model Card},
  author={{Qwen Team}},
  year={2026},
  howpublished={\url{https://huggingface.co/Qwen/Qwen3.6-27B}},
  note={Accessed July 22, 2026}
}

@misc{qwen330ba3b,
  title={{Qwen3-30B-A3B} Model Card},
  author={{Qwen Team}},
  year={2025},
  howpublished={\url{https://huggingface.co/Qwen/Qwen3-30B-A3B}},
  note={Accessed July 31, 2026}
}

@misc{google2026gemma431b,
  title={{Gemma 4 31B} Model Card},
  author={{Google DeepMind}},
  year={2026},
  howpublished={\url{https://huggingface.co/google/gemma-4-31B}},
  note={Accessed July 31, 2026}
}

@article{jiang2026sok,
  title={{SoK}: Agentic Skills -- Beyond Tool Use in {LLM} Agents},
  author={Jiang, Yanna and Li, Delong and Deng, Haiyu and Ma, Baihe and Wang, Xu and Wang, Qin and Yu, Guangsheng},
  journal={arXiv preprint arXiv:2602.20867},
  year={2026}
}

@article{liu2025structured,
  title={On the Structural Memory of {LLM} Agents},
  author={Zeng, Ruihong and Fang, Jinyuan and Liu, Siwei and Meng, Zaiqiao},
  journal={arXiv preprint arXiv:2412.15266},
  year={2024}
}

\end{document}